\documentclass{ecai} 

\usepackage{multicol,multirow}
\usepackage{array}
\usepackage{makecell}
\usepackage{color}
\usepackage[table,xcdraw]{xcolor}
\definecolor{ForestGreen}{rgb}{0.13, 0.55, 0.13}  

\usepackage[hang,flushmargin]{footmisc}

\newcommand{\ie}{i.e.}  

\usepackage{colortbl}  
\usepackage{xcolor}
\usepackage{array}
\usepackage{arydshln}
\definecolor{gg}{HTML}{e2f0cb}

\usepackage{latexsym}
\usepackage{amssymb}
\usepackage{amsmath}
\usepackage{amsthm}
\usepackage{booktabs}
\usepackage{enumitem}
\usepackage{graphicx}
\usepackage{color}

\newcommand{\BibTeX}{B\kern-.05em{\sc i\kern-.025em b}\kern-.08em\TeX}

\begin{document}


\begin{frontmatter}


\paperid{1532} 


\title{LoCa: Logit Calibration for Knowledge Distillation}


\author[1]{\fnms{Runming}~\snm{Yang}}
\author[2]{\fnms{Taiqiang}~\snm{Wu}}
\author[1]{\fnms{Yujiu}~\snm{Yang}\thanks{Corresponding Author. Email: yang.yujiu@sz.tsinghua.edu.cn}} 

\address[1]{Tsinghua University}
\address[2]{The University of Hong Kong}


\begin{abstract}
Knowledge Distillation~(KD), aiming to train a better student model by mimicking the teacher model, plays an important role in model compression. 
One typical way is to align the output logits.
However, we find a common issue named mis-instruction, that the student would be misled when the predictions based on teacher logits do not follow the labels.
Meanwhile, there is other useful dark knowledge in the logits such as the class discriminability, which is vital for distillation.
In this paper, we propose a simple yet effective \textbf{Lo}git \textbf{Ca}libration~(LoCa) method, which calibrates the logits from the teacher model based on the ground-truth labels.
The key insight is to correct the prediction~(to address the mis-instruction issue) and maintain useful dark knowledge simultaneously.
Our proposed LoCa does not require any additional parameters.
Empirical results on image classification and text generation tasks demonstrate that LoCa can effectively improve the performance of baselines.
\end{abstract}

\end{frontmatter}

\section{Introduction}

In the last decade, the development of deep neural networks has revolutionized the field of computer vision~(CV)~\citep{he2016deep,senet,shufflenetv2} and natural language processing~(NLP)~\citep{DBLP:journals/corr/abs-1907-11692}. 
Complex network architectures~\cite{shelhamer2016fully,pspnet} and increasing parameter~\citep{DBLP:conf/naacl/DevlinCLT19} can make a stronger model but also bring high costs in computation and deployment~\citep{faster_rcnn,mask}.
Such costs are not preferable when applying models to industrial scenarios, and thus researchers have made many efforts to compress the models.
One mainstream approach to designing lightweight models is knowledge distillation~(KD)~\citep{hinton2015distilling, wu2024weight}, which concentrates on transferring the knowledge from a heavy model~(i.e. teacher) to a light one~(i.e. student).

The goal of KD is to train a better student by mimicking the teacher~\citep{lao2023masked,lao2023unikd}.
For example, the logit-based KD \citep{hinton2015distilling} employs the KL divergence to align the logits for classification.
Compared to the one-hot labels that only contain the target category information, the logits encompass predictive information for all categories, which is also known as dark knowledge.
In this way, we can learn a better student with such dark knowledge.

\begin{figure}[!t]
\includegraphics[width=0.5\textwidth]{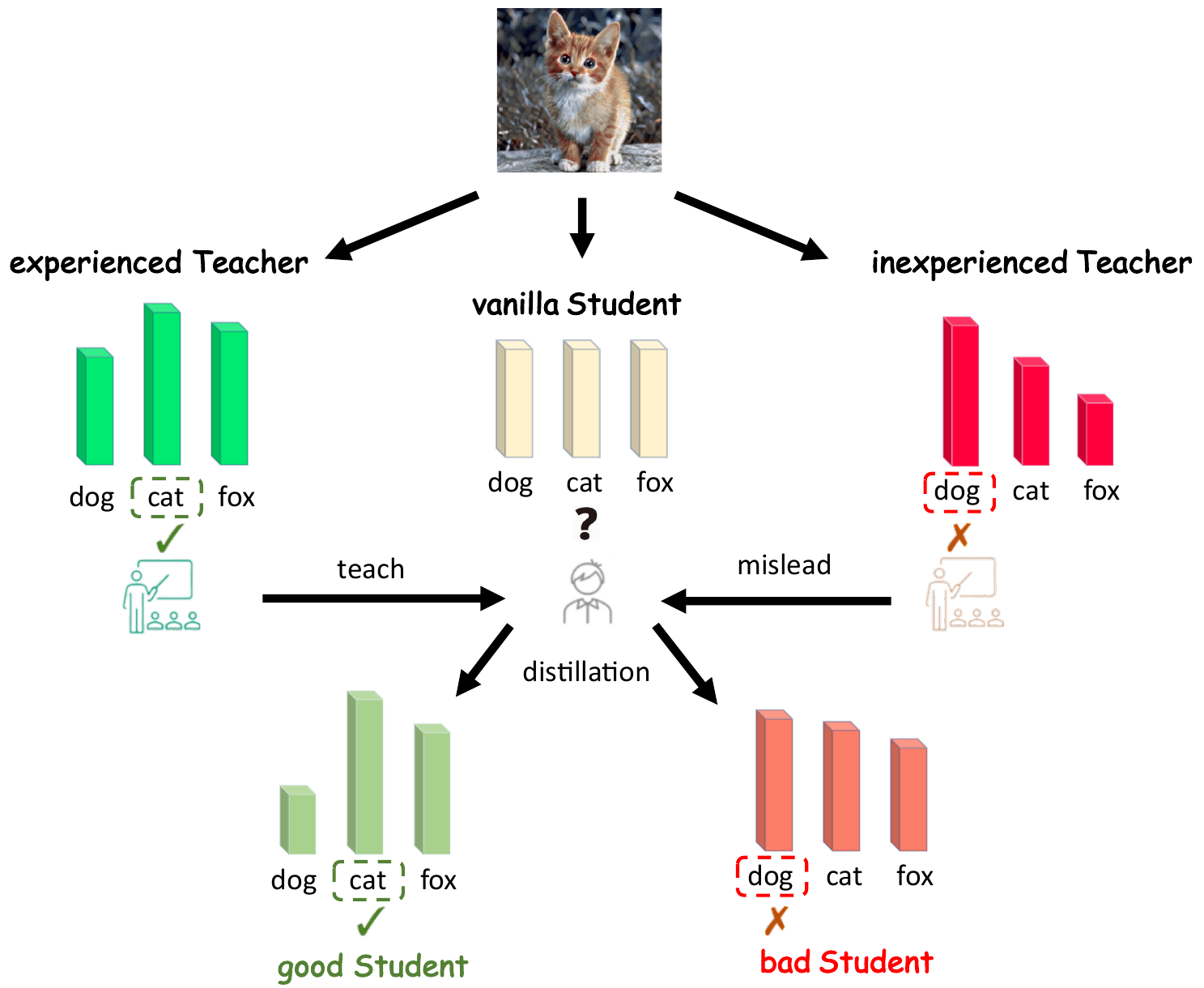} 
\caption{
One example of the \textbf{mis-instruction} issue.
The student model can generate the right prediction under an experienced teacher ~(in \textcolor[RGB]{0,180,0}{green}), but would be misled under the inexperienced teacher~(in \textcolor[RGB]{180,0,0}{red}).
}
\label{case}
\end{figure}

However, we argue that there exists an issue named \textbf{mis-instruction}, where the student would be misled when the teacher logits are wrong.
Specifically, when the predictions based on teacher logits do not follow the labels, such erroneous would mislead the student model in the knowledge distillation process.
Figure \ref{case} shows one example of the mis-instruction issue where the input is an image of a cat.
Under the experienced teacher with the right predictions~(left side), the student model can effectively learn such knowledge and classify the image as the 'cat'.
In contrast, when the logits from the teacher model contain the wrong prediction~(a.k.a the inexperienced teacher on the right), the student model would be misled and generate a wrong prediction 'dog'.

\begin{figure}[!t]
\includegraphics[width=0.46\textwidth]{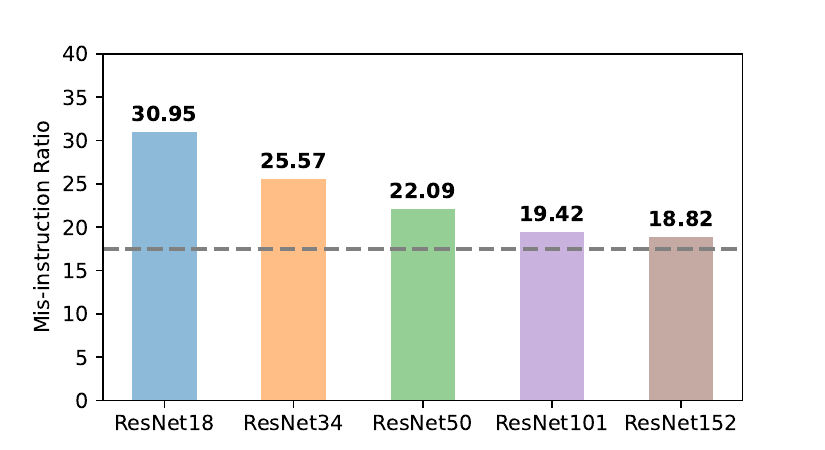} 
\caption{
Mis-Instruction ratio of various teacher models on ImageNet training set.
For all models, the ratios are greater than \textbf{17.5\%}.
}
\label{mis-intruction ratio}
\end{figure}

Meanwhile, the mis-instruction phenomenon is common in practice.
As shown in Figure \ref{mis-intruction ratio}, the mis-instruction ratios on the ImageNet training set are notably high.
Particularly, for the ResNet101, a typical teacher model on ImageNet, there are as many as 19.4\% of the samples where the predictions based on teacher logits do not follow the labels.

To address the mis-instruction issue, one intuitive idea is to skip the samples with wrong logits in distillation.
However, such logits also contain other valuable dark knowledge~\citep{chandrasegaran2022revisiting}, such as the class discriminability.
The class discriminability refers to the ratios of the logit for all the non-target classes and is vital in the distillation process \citep{li2022asymmetric}.
Simply discarding these logits would also lead to the loss of such valuable dark knowledge. 
We thus ask: \textbf{can we calibrate the logits to both avoid mis-instruction and maintain other useful dark knowledge?}

In this paper, we propose a novel Logit Calibration~(LoCa) method to calibrate the logits from the inexperienced teacher without any additional parameters.
Our key insight is to guarantee that teacher logits are consistent with the ground-truth label~(to avoid mis-instruction) and also maintain the ratio of the non-target logits~(to maintain the useful dark knowledge).
Specifically, we model the calibration process as an optimization problem and propose a feasible solution by introducing a scaling factor. 
This optimization problem modeling revolves around three perspectives, namely 1) probability distribution, 2) prediction correctness, and 3) non-target proportion invariance.
After that, we employ the calibrated logits in the knowledge distillation.

We perform experiments on 1) image classification tasks on CIFAR-100~\citep{cifar} and ImageNet~\citep{imagenet}, and 2) text generation tasks on Dolly~\citep{DBLP:journals/corr/abs-2306-08543}, S-NI~\citep{wang-etal-2022-super} and UnNI~\citep{honovich-etal-2023-unnatural} datasets. 
Experimental results indicate that our proposed LoCa significantly outperforms the baselines, demonstrating the effectiveness of calibrating logits.
Moreover, further analysis shows that within the hyperparameter alpha range of 0.9 to 1.0, LoCa exhibits high usability and robustness.
In conclusion, the main contributions are as follows:
\begin{itemize}
    \item We find an issue termed mis-instruction, where the student model would be misled when the predictions based on teacher logits do not follow the labels.
    \item We propose a simple yet effective strategy, LoCa, which calibrates the teacher logits to avoid mis-instruction and maintain other useful dark knowledge.
    \item We conduct experiments on image classification and text generation tasks. The results of the experiment demonstrate the effectiveness of the proposed LoCa method. 
\end{itemize}

\section{Preliminary}
\label{prelimi}

In this section, we introduce the fundamental principles of knowledge distillation (KD), specifically focusing on the logit-based KD approach. 
After that, we also explore the mis-instruction issue and investigate it by analyzing the definition of its samples, demonstrating its detrimental effects through a comparative experiment.

\subsection{Logit-based KD}

The vanilla standard KD~\citep{hinton2015distilling} transfers knowledge by aligning the output logits of the teacher and student models. 
Considering a classification task with $C$ classes, 
the logits from the teacher model $\mathbf{p}$ can be formulated as follows:
\begin{equation}
    \mathbf{p} = [p_1, p_2, ..., p_C] \in \mathbb{R}^{1 \times C},
    \label{teacher_p}
\end{equation}
where $p_i$ is the probability of the $i$-th class.
Typically, $p_i$ is obtained through the softmax function with temperature:
\begin{equation}
    p_i = \frac{\exp(z_i/\tau)}{\sum_{j=1}^{C} \exp(z_j/\tau)},
    \label{softmax}
\end{equation}
where $z_i$ represents the logit of $i$-th class and $\tau$ is the temperature parameter for scaling.

Similarly, the output logits $\mathbf{q}$ from the student model can be denoted as follows:
\begin{equation}
    \mathbf{q} = [q_1, q_2,..., q_C] \in \mathbb{R}^{1 \times C},
    \label{student_q}
\end{equation}
where $q_i$ is the predicted probability for the $i$-th class.

To align the logits, we can employ the Kullback-Leibler~(KL) divergence, which can be written as: 
\begin{equation}
    \text{KD}(\mathbf{p}, \mathbf{q}) = \sum\limits_{i=1}^{n} p_i \log\left(\frac{p_i}{q_i}\right).
    \label{eq:kd_loss}
\end{equation}

Another goal is to generate the right prediction for the student model, where the cross-entropy loss is widely used.
Specifically, the cross-entropy loss measures the discrepancy between the predicted probabilities $\mathbf{q}$ and the one-hot labels $\mathbf{y} \in \mathbb{R}^{1\times C}$ as follows:
\begin{equation}
    \text{CE}(\mathbf{q}, \mathbf{y}) = \sum\limits_{i=1}^{n} y_i \log(q_i) = \log(q_{gt}).
\label{ce_loss}
\end{equation}
where ${gt}$ denotes the ground truth class and $\mathbf{y}=[y_1,...,y_C]$ is defined as:
\begin{equation}
y_i= \begin{cases}
\: 1 \ \quad \text{if} \ i \ \text{is} \ {gt} \\ 
\: 0 \ \quad \text{otherwise}.
\end{cases}
\label{one-hot}
\end{equation}

Therefore, we can get the final optimal object in knowledge distillation:
\begin{equation}
    \mathcal{L} = \beta \cdot \text{KD}(\mathbf{p}, \mathbf{q}) + \gamma \cdot \text{CE}(\mathbf{q}, \mathbf{y}),
\label{eq:Loss function}
\end{equation}
where $\beta$ and $\gamma$ are hyperparameters that can be respectively tuned for optimal training performance. We adopted fixed parameters based on previous studies for fair comparisons. See Section \ref{Sec:Implementation} for detailed experimental settings.

\begin{table}[!t]
\centering
\caption{Results on CIFAR-100 dataset.
* denotes the strategy that dropping the mis-instruction samples during the distillation process.}
\begin{tabular}{lcc}
\toprule
\textbf{Model}   & \textbf{ResNet56 $\rightarrow$ResNet20} & \textbf{WRN-40-2$\rightarrow$WRN-40-1} \\
\midrule
Teacher &   72.34       &   75.61      \\
Student &     69.06     &   70.50     \\
KD      &  70.66        &  73.54        \\
KD*     & 70.88         &     73.59   \\
\bottomrule
\end{tabular}

\label{kd_mis}
\end{table}

\begin{figure*}[!t]
\includegraphics[width=1\textwidth]{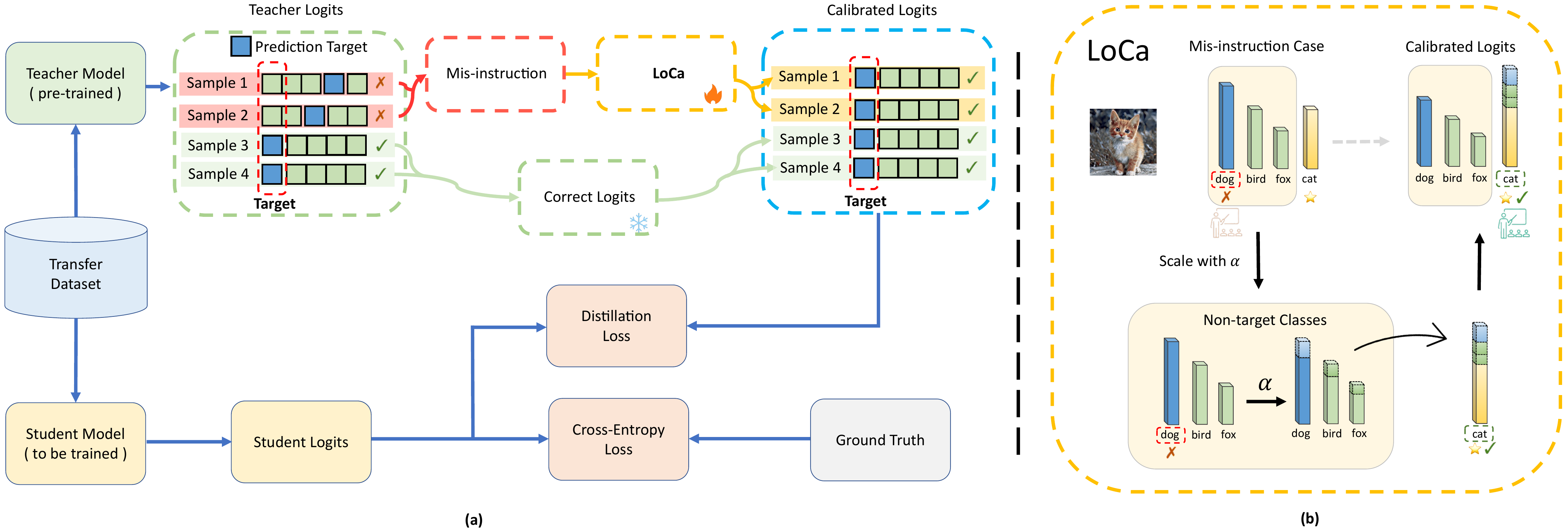} 
\vspace{-15pt}
\caption{
The details of the proposed LoCa method.
During distillation, we first calibrate the logits for the mis-instruction samples and then employ the calibrated logits for KD.
Specifically, we introduce a scaling factor $\alpha$ to decrease the non-target logits and increase the target logit. 
}
\label{pipeline}
\end{figure*}

\subsection{Analysis on Mis-instruction Issue}
\label{Analysis on Mis-instruction}

Mis-instruction occurs when the logits of teacher model $\mathbf{p}$ indicate wrong class labels, transferring incorrect logits information to the student model $\mathbf{q}$. 
We first find the highest logits class index $k_{\text{logits}}$ as follows:
\begin{equation}
k_{\text{logits}}=\text{Argmax}_i\{p_i\}.
\end{equation}
After that, the mis-instruction samples are defined as:
\begin{equation}
    k_{\text{logits}} \neq {gt},
    \label{wrong samples}
\end{equation}
where ${gt}$ represents the index of the ground truth class. 

Theoretically, mis-instruction samples would mislead students during the distillation.
Specifically, the cross-entropy loss aims to optimize student logits toward ground truth label ${gt}$, while the KD loss aims to optimize it toward the label $k_{\text{logits}}$ indicated by teacher logits. 
As shown in Equation \ref{wrong samples}, these two goals are not the same, leading to an optimization conflict.

Moreover, we perform experiments on a straightforward strategy to skip these mis-instruction samples.
As shown in Table \ref{kd_mis}, the student model ResNet20 trained without mis-instruction samples gets a higher result of 70.88 than 70.66 of the vanilla KD, with other conditions being completely consistent.
Thus, we can conclude that the mis-instruction issue exists and that it is \textbf{harmful to students} during the distillation process.

\section{Methodology}

In this section, we introduce the motivations and details of our proposed LoCa method to address the mis-instruction issue.

\subsection{Optimization Objective}
LoCa is designed under two goals: 1) addressing the mis-instruction issue, and 2) preserving the proportion of non-target classes, which aims to maintain the valuable dark knowledge information. 
Following \citet{zhao2022decoupled}, we divide the logits into two parts, i.e.,  non-target and target categories.
Then we model the calibration goals as an optimization problem.

\noindent \textbf{Probability distribution.}
As a set of probability distributions, the calibrated logits $\mathbf{p}^{\text{loca}} = [p_1^{loca}, p_2^{loca}, ..., p_C^{loca}] \in \mathbb{R}^{1 \times C}$ needs to satisfy the constraint condition that the sum of probabilities equals 1, which can be formulated as:

\begin{equation}
    \sum_{i=1}^{C} p_i^{loca} = 1
    \label{eq:probability distribution}
\end{equation}
where each element represents a valid probability with
\begin{equation}
      0 < p_i^{loca} < 1 \quad \forall i=1,2,3,..,C \ .
\end{equation}

\noindent \textbf{Prediction correctness.} 
To address the mis-instruction issue, the goal is to guarantee the predicted label to be consistent with the ground truth label, which can be formulated as:

\begin{equation}
    k_{\text{logits}}^{loca} = \text{Argmax}_i\{p_i^{loca}\} = {gt}.
    \label{eq:prediction correct}
\end{equation}

\noindent \textbf{Non-target proportion invariance.}
As shown in Figure \ref{pipeline}, we separate the logits into target and non-target categories.
The next goal is to maintain useful dark knowledge in the non-target logits \citep{zhao2022decoupled}.
However, simply dropping these mis-instruction samples would also bring the loss of vital dark knowledge.
To preserve this knowledge, the key is to maintain the ratios between any two non-target logits:
\begin{equation}
    \frac{p_i^{loca}}{p_j^{loca}} = \frac{p_i}{p_j} \quad \forall i, j \neq {gt}.
    \label{eq:non-target in-variance}
\end{equation}

\begin{table*}[!t]
\centering
\caption{Results of various methods on the CIFAR-100 validation dataset. Results are averaged over 3 trials.
}
\setlength{\belowcaptionskip}{-10pt}
\begin{tabular}{cccccccccccc}
\toprule
 \multirow{2}{*}{Teacher}  & ResNet56 & ResNet110 & ResNet32$\times$4  & WRN-40-2& WRN-40-2   & WRN-40-2       & ResNet50      & ResNet32$\times$4 \\
                            & 72.34      & 74.31       & 79.42         & 75.61   &   75.61       & 75.61          & 79.34      & 79.42  \\
\multirow{2}{*}{Student}  & ResNet20 & ResNet32 & ResNet8$\times$4   & WRN-16-2 & WRN-40-1 & ShuffleNet-V1   & MobileNet-V2   & ShuffleNet-V2  \\
                            & 69.06      & 71.14       & 72.50         & 73.26  &   71.36        & 70.50         & 64.60      & 71.82  \\ 
\midrule
KD \cite{hinton2015distilling}   & 70.66 & 73.08 & 73.33                & 74.92& 73.54          & 74.83            & 67.35      & 74.45         \\
 \textbf{LoCa-0.95}          & \textbf{71.08}&\textbf{73.36}&{73.66}  &{75.11}&{73.74}          &{75.42}        &\textbf{68.66}   & \textbf{75.30} \\
  $\Delta$      & \textcolor{ForestGreen}{+0.42}   & \textcolor{ForestGreen}{+0.28}  & \textcolor{ForestGreen}{+0.33}        
                                            & \textcolor{ForestGreen}{+0.19}   & \textcolor{ForestGreen}{+0.20} 
                                                                        & \textcolor{ForestGreen}{+0.59}    & \textcolor{ForestGreen}{+1.31}& \textcolor{ForestGreen}{+0.85}
\\
  \textbf{LoCa-0.98}        & {70.88} & {73.32} & \textbf{73.79}    &\textbf{75.21}&\textbf{73.85} & \textbf{75.81}   & 68.60        & 75.10 \\
 $\Delta$      & \textcolor{ForestGreen}{+0.22}      & \textcolor{ForestGreen}{+0.24}       & \textcolor{ForestGreen}{+0.46}        
                                            & \textcolor{ForestGreen}{+0.29}      & \textcolor{ForestGreen}{+0.31} 
                                                                                    & \textcolor{ForestGreen}{+0.98}& \textcolor{ForestGreen}{+1.25}& \textcolor{ForestGreen}{+0.65}
\\
\bottomrule
\end{tabular}
\label{tab:CIFAR100}
\end{table*}

\subsection{LoCa: Logit Calibration}
In this paper, we design the \textbf{Lo}git \textbf{Ca}libration~(LoCa) method to achieve the aforementioned goals.
Figure \ref{pipeline} shows the details.

For the mis-instruction sample, the calibrated logit is defined as:
\begin{equation}
    p_i^{loca} = 
    \begin{cases} 
        \; s \cdot p_i & \text{if } i \neq {gt}, \\
        \; 1 - \sum\limits_{i=1, i\neq k}^{C} p_i^{loca} & \text{if } i = {gt},
    \end{cases}
    \label{eq:loca_p}
\end{equation}

Considering the prediction correctness shown in Equation \ref{eq:prediction correct}, the logit at the ground truth label $gt$ should be \textbf{only} maximum among these logits:
\begin{equation}
  p_{gt}^{loca} > p_{i}^{loca} \ \ \ \forall i \neq gt,
\end{equation}
which equals
\begin{equation}
  p_{gt}^{loca} > \max\limits_{i\neq gt}(p_{i}^{loca}) =   p_{k_{\text{logits}}}^{loca}.
  \label{eq:inequality}
\end{equation}

Combining the Equation ~\ref{eq:loca_p} and \ref{eq:inequality}, we have:
\begin{equation}
    1 - s \cdot \sum_{i \neq k} p_i > s \cdot p_{k_{\text{logits}}}.
\end{equation}
Therefore, we can get a feasible solution: 
\begin{equation}
    s <  \sigma = \frac{1}{1 - p_{gt} + p_{k_{\text{logits}}}} ,
\end{equation}
where the $\sigma$ is the threshold.
Thus, we introduce a hyperparameter \( \alpha \) ranging between 0 and 1 and set \( s = \alpha \cdot \sigma \). 
This configuration ensures that the LoCa method satisfies the aforementioned constraints. 
We then scale the logits in both target and non-target categories accordingly based on Equation \ref{eq:loca_p}.

\section{Experiments}
In this section, 
we perform experiments on two representative tasks in CV and NLP fields, \ie, image classification, and text generation.

\subsection{Image Classification Tasks}

\subsubsection{Datasets and Models}
\noindent{\textbf{Datasets. }
{{CIFAR-100}~\cite{cifar}} is a well-known image classification dataset, containing $32\times 32$ images of 100 categories. 
Training and validation sets contain 50,000 and 10,000 images.
ImageNet~\cite{imagenet} is a large-scale classification dataset that consists of 1000 classes. The training set contains 1.28 million images and the validation set contains 50,000 images.

\noindent{\textbf{Model.}}
We employ different ResNet architectures as teacher models, including ResNet56, ResNet110, ResNet32×4, and WRN-40-2. For student models, we select both homologous ResNet structures and heterologous ShuffleNets and MobileNets.

\subsubsection{Baselines and Implementation}\label{Sec:Implementation}

\noindent\textbf{Baselines.} We primarily test the enhancements and performance improvements of our strategy on the vanilla KD~\cite{hinton2015distilling} approach.

\noindent\textbf{Implementation.} We follow the same experimental settings as in previous work~\cite{chen2021distilling,zhao2022decoupled}. 
For the experiments on CIFAR-100, the optimizer is SGD~\cite{sutskever2013importance} and trained for $240$ epochs. 
The learning rate is initialized as 0.01 for MobileNets~\cite{howard2017mobilenets,sandler2018mobilenetv2} and ShuffleNets~\cite{zhang2018shufflenet}, and 0.05 for ResNets~\cite{he2016deep} and WRNs~\cite{zagoruyko2016wide}. 

For a fair comparison, we fix $\tau$ from Equation \ref{softmax} and $\beta$, $\gamma$ from Equation \ref{eq:Loss function} across experiments: $\tau=4$, $\beta=0.9$, $\gamma=0.1$ for CIFAR-100 and $\tau=1$, $\beta=0.5$, $\gamma=0.5$ for ImageNet.
We report the average results over 3 trials.
It takes around 2 hours to train on 1 Nvidia A100 GPU for CIFAR-100 and around 24 hours for ImageNet.

\subsubsection{Main Results}

\noindent\textbf{Results on CIFAR-100.} 
Table~\ref{tab:CIFAR100} reports the validation accuracy. 
Based on the results, we can get the following findings:

\begin{itemize}

    \item Our proposed LoCa can consistently improve the performance of distillation compared to baseline vanilla KD~\cite{hinton2015distilling}. 
    In particular, LoCa achieves 71.08\% with ResNet56 as teacher and ResNet20 as student, which is 0.42 higher than the original KD method.

    \item We find that LoCa improves distillation performance for both homologous teacher-student pairs and heterologous pairs such as ResNet50 to MobileNet-V2, demonstrating the robustness towards model structures.

    \item We observe that different values of $\alpha$ would lead to variations in the benefits of our LoCa strategy, yet the overall trend remains an improvement, showing the robustness towards different hyperparameters.

\end{itemize}

\begin{figure*}[!t]
\centering
\includegraphics[width=0.95\textwidth]{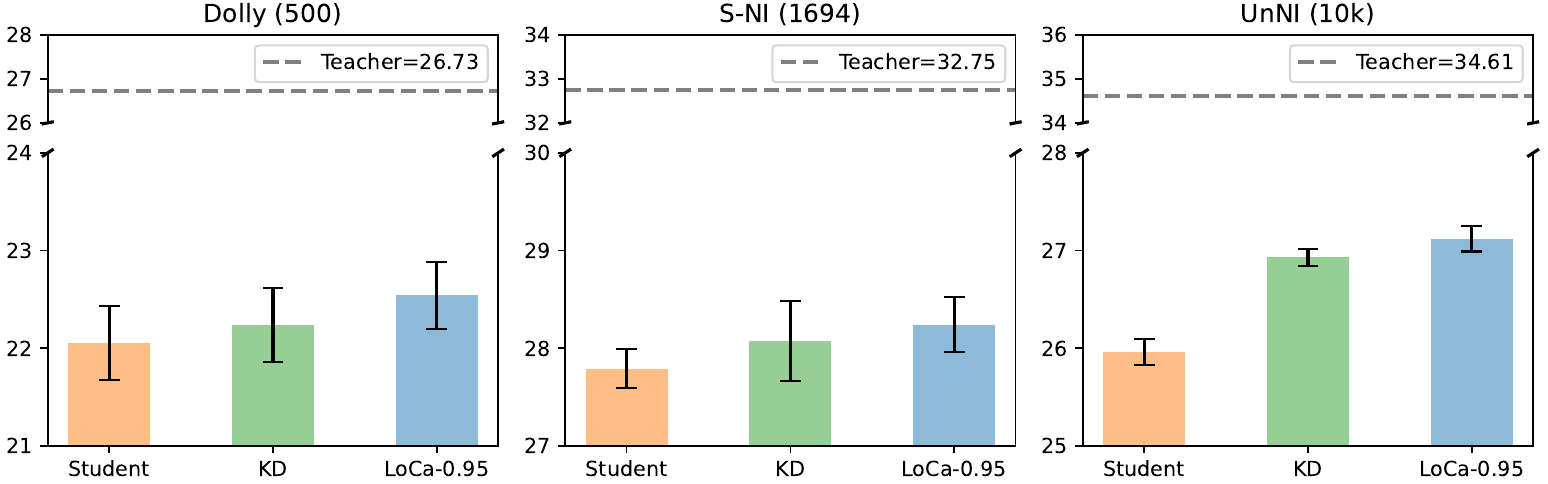} 
\caption{
Rouge-L scores on Dolly, S-NI, and UnNI datasets.
We report the average and standard deviation scores for 5 trials.
Our proposed LoCa outperforms KD on all benchmarks.
}
\label{llm_result}
\end{figure*}

\begin{table}[!t]
\center
\caption{Top-1 and top-5 accuracy (\%) on the ImageNet validation from \textbf{ResNet-34} to \textbf{ResNet-18}.
The results are averaged over 3 trials.}
\vspace{1.8em}
\resizebox{\linewidth}{!}{
\begin{tabular}{cccccc}
\toprule
Metrics & Teacher & Student &KD & LoCa-0.95 & LoCa-0.99 \\
\midrule
top-1 & 73.31 & 69.75 &70.66 & 71.08 & 71.15\textcolor{ForestGreen}{~(+0.49)}\\
top-5 & 91.42 & 89.07 &89.88 & 90.09 & 90.19\textcolor{ForestGreen}{~(+0.31)}\\
\bottomrule
\end{tabular}
}
\label{tab:ImageNet_res34_res18}
\vspace{1em}
\end{table}

\begin{table}[!t]
\center
\caption{Top-1 and top-5 accuracy (\%) on the ImageNet validation from \textbf{ResNet-50} to \textbf{MobileNetV1}.
The results are averaged over 3 trials.}
\vspace{1.8em}
\resizebox{\linewidth}{!}{
\begin{tabular}{cccccc}
\toprule
Metrics & Teacher & Student & KD & LoCa-0.95 & LoCa-0.99 \\
\midrule
top-1 & 76.16 & 68.87 & 70.50 & 70.91 & 70.99\textcolor{ForestGreen}{~(+0.49)}\ \\
top-5 & 92.86 & 88.76 & 89.80 & 90.03 & 90.06\textcolor{ForestGreen}{~(+0.26)}\ \\
\bottomrule
\end{tabular}
}
\vspace{1em}
\label{tab:ImageNet_res50_mobilev1}
\end{table}

\noindent\textbf{Results on ImageNet.} 
We report the top-1 and top-5 accuracies of image classification on ImageNet in Table~\ref{tab:ImageNet_res34_res18} and Table~\ref{tab:ImageNet_res50_mobilev1}.
Similarly, we find that our LoCa can achieve a consistent improvement in top-1 and top-5 accuracy on ImageNet validation. 
Specifically, LoCa gets an improvement of 0.49\% on the top-1 accuracy under the setting from ResNet-34 to ResNet-18.
Also, LoCa can improve the performance ranging from different settings.
The findings are consistent with the CIFAR, which shows the robustness of the proposed LoCa towards various benchmarks.

\subsection{Text Generation Tasks}

We follow the same experimental settings as~\citet {wu2024rethinking}, first fine-tuning the teacher model and then distilling the teacher model.
After that, we report the average Rouge-L scores on popular benchmarks.

\subsubsection{Datasets and Models}

\noindent\textbf{Datasets.}
For training data, we employ the instruction response dataset following \citet{DBLP:journals/corr/abs-2306-08543}, which is built from databricks-dolly-15k\footnote{https://github.com/databrickslabs/dolly/tree/master} and contains 14k samples for training, 500 samples for valid, and 500 samples for testing.

The details for the evaluation dataset are as follows:
\begin{itemize}
    \item Dolly: human-written instruction-response pairs 
    We divide the data set into 14k samples for training, 500 samples for validation, and 500 samples for testing following \citet{DBLP:journals/corr/abs-2306-08543}.
    \item S-NI: the test set of SUP-NATINST \citep{wang-etal-2022-super}, which contains 9K samples from 119 English tasks.
    In this paper, we employ the samples with ground truth responses longer than 11.
    \item UnNI: dataset from \citet{honovich-etal-2023-unnatural}.
    Similarly, we employ samples with ground-truth responses longer than 11.
\end{itemize}

\noindent\textbf{Models.}
We perform experiments on the mainstreaming model LLaMA \citep{DBLP:journals/corr/abs-2302-13971}.
Specifically, we employ LLaMA with 7B parameters as our teacher model and TinyLLaMA \citep{zhang2024tinyllama} with 1.1B parameters\footnote{https://huggingface.co/TinyLlama/TinyLlama-1.1B-intermediate-step-1195k-token-2.5T} as the student model. This allows us to evaluate the effectiveness of LoCa across widely recognized model architectures.

\subsubsection{Baselines and Implementation}
\noindent\textbf{Baseline.}
For the baseline student model, we directly train the student model by performing SFT on the dataset without any distillation and denote it as "Student".
Moreover, we employ {SeqKD} \citep{kim2016sequence} to train the student model on the data generated from the teacher model, which we denote as "KD" for the consistency of the expression.

\noindent\textbf{Implementation.}
For TinyLLaMA, we set the batch size as 60 and train for 10 epochs.
The learning rate is 1e-5.
For the student, the maximum input length is 512.
It takes around 1h to train on 4 Nvidia A100 GPUs.
We report the results of the Rouge-L scores for five different seeds.

\begin{table}[!t]
\center
\caption{Statistics of the datasets for distillation on the text generation tasks.}
\vspace{1.5em}
\begin{tabular}{ccccc}
\toprule
 \textbf{Dataset Name} & \textbf{Usage} & \textbf{Samples} \\
\midrule
 \multirow{2}{*} &{Train} & 14,000 \\
\cmidrule{2-3}
{Dolly} & {Valid} & 500 \\
\cmidrule{2-3}
 & {Test} & 500 \\
\midrule
 {S-NI} & Test & 1,694 \\
\cmidrule{1-3}
 {UnNI} & Test & 10,000 \\
\bottomrule
\end{tabular}
\label{tab:dataset_split}
\vspace{2em}
\end{table}

\begin{figure*}[!t]
\centering
\includegraphics[width=1\textwidth]{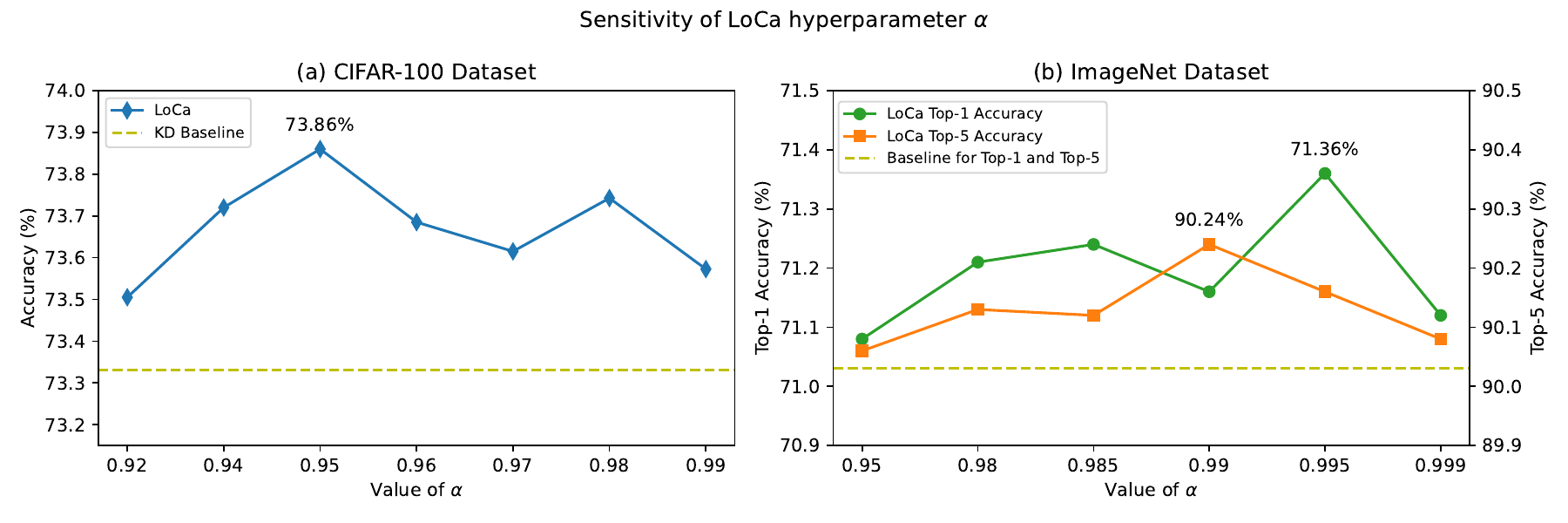} 
\vspace{-5pt}
\caption{
The ablation studies under different $\alpha$ settings in our LoCa.
We employ \textbf{ResNet32$\times$4} and \textbf{ResNet8$\times$4} as the teacher and the student on CIFAR-100~(part a).
We set \textbf{ResNet-34} to \textbf{ResNet-18} as the teacher and student on ImageNet~(part b).
}
\label{ablation_result}
\vspace{1em}
\end{figure*}

\subsubsection{Main Results}
As shown in Figure~\ref{llm_result}, we present the average Rouge-L scores and variance for students trained using SFT, KD, and LoCa methods on the Dolly, S-NI, and UnNI datasets. 
The dashed lines indicate the performance of the teacher model. 
We can observe consistent improvements employing our LoCa method across three datasets, compared to both the students trained via SFT and those transferred via KD, which demonstrates the effectiveness of our method. 
In particular, on the larger UnNI dataset with 10,000 samples, LoCa outperforms both SFT and KD, with p-values less than 0.001 for both comparisons, indicating a statistically significant improvement.
Notably, on the Dolly dataset, the enhancements from our LoCa method to KD surpass the improvements from SFT to KD. 
Furthermore, the variance of LoCa is much smaller than the baselines, which shows the effectiveness of distilling knowledge.

\section{Analysis}
In this section, we perform further detailed analyses on LoCa and apply LoCa on more baselines such as DKD.

\subsection{Impact of the  $\alpha$}
\label{sec:ablation}

We perform ablation studies to assess the sensitivity of our approach to the hyperparameter $\alpha$.
Figure~\ref{ablation_result} reports the student accuracy~(\%) with different $\alpha$, where we employ {ResNet32$\times$4} and {ResNet8$\times$4} as the teacher and student on CIFAR-100~(left), ResNet32 and ResNet18 on ImageNet~(right). 
We can observe that the improvements on CIFAR-100 are relatively stable, with $\alpha=0.95$ achieving the peak performance within the ablation range.
Meanwhile, the performance on ImageNet is also improved, proving that LoCa can consistently increase model performance, especially when $\alpha=0.995$. 
Moreover, such consistency over different hyperparameters indicates the robustness of our LoCa in achieving consistent improvements under different settings.

Meanwhile, we find that the effects on ImageNet are more sensitive compared to CIFAR-100, specifically showing that when $\alpha=0.95$ or $\alpha=0.999$, there are generally slight improvements or performances close to the original. 
We attribute the difference in performance between ImageNet and CIFAR-100 to the larger size and the larger number of categories in the ImageNet dataset, suggesting that the adjustments must be moderate. 



We also perform experiments when $\alpha$ is equal to or greater than 1.
In this case, the predicted labels are not guaranteed to be the ground truth.
As shown in Table \ref{tab:ablation_alpha2}, we can find that LoCa performs worse than the baseline vanilla KD, indicating the importance of prediction correctness.

\begin{table}[!t]
\centering
\caption{Sensitivity of $\alpha$ around 1.
We report the Top-1 accuracy (\%) from \textbf{ResNet34} to \textbf{ResNet18} on ImageNet.
When $\alpha$ is equal or larger than 1, the performance would be worse than the baseline.
}
\begin{tabular}{cccccc}
\toprule
\multirow{2}{*}{\textbf{Metric}} & \multirow{2}{*}{\textbf{Vanilla KD}} & \multicolumn{4}{c}{\textbf{The value of $\alpha$ in LoCa}} \\
\cmidrule{3-6}
&& {0.995} & {0.999} &\textbf{1.000} & \textbf{1.010} \\
\midrule
Top-1  & \textbf{71.03} & \textcolor{ForestGreen}{71.36} & \textcolor{ForestGreen}{71.12} &\textcolor{red}{71.01}&\textcolor{red}{70.94} \\
\bottomrule
\end{tabular}
\label{tab:ablation_alpha2}
\vspace{2em}
\end{table}

\subsection{Time Cost}

In assessing the additional timing introduced by the LoCa method during the distillation on CIFAR-100, LoCa incurs a minor computational cost, as detailed in Table~\ref{tab:function_timing}. 
Specifically, from WRN\_40\_2 to ShuffleV1, the KD Loss computation time increased from 0.32 ms to 0.39 ms, indicating a marginal additional computational overhead of 0.07 ms by LoCa. 
Furthermore, despite frequent function calls (i.e., 781 times per epoch), the additional time per call is less than 0.1 ms, contributing less than 0.1 seconds to the approximately 28-second duration of each epoch.

This slight proportional increase is consistent with our anticipated increases in computational cost, suggesting that the LoCa method improves performance \textbf{without requiring significant additional time}~(less than 1\%).

\begin{table}[!t]
\centering
\caption{Time costs for LoCa and baselines.
We report the average scores for five trials.
KD Loss denotes the time costs for calculating KL divergence.
Batch denotes the time costs for one batch~(64 samples) and Epoch for the process of training one epoch.
}
\begin{tabular}{ccccc}
\toprule
\textbf{Method}&\textbf{KD Loss}&\textbf{Batch}&\textbf{Epoch}\\
\midrule
KD & 0.32 ms & 27.83 ms & 25.32 s \\
\cmidrule{1-5}
LoCa & 0.39 ms & 28.29 ms  & 25.57 s \\
\cmidrule{1-5}
$\Delta$  & 21.88\% & 1.65\%  & 0.99\% \\
\bottomrule
\end{tabular}
\label{tab:function_timing}
\end{table}


\subsection{LoCa on DKD}

We conducted extensive experiments on the KD baseline, and the results demonstrate the effectiveness and robustness of the proposed LoCa.
Nevertheless, LoCa can be easily extended on more baselines, such as DKD \citep{zhao2022decoupled}.
Specifically, we first apply LoCa to adjust the logits on the mis-instruction samples, followed by the standard process of DKD.
Table \ref{loca_dkd} shows the results of the LoCa with various $\alpha$.
We report the average score of 3 trials.
We can find that applying LoCa would improve the performance of DKD in all $\alpha$, demonstrating the effectiveness.
Specifically, DKD with LoCa ($\alpha$=0.95) would get 77.25, 0.37 higher than the vanilla DKD, when distilling ResNet32x4 to ShuffleNet-V2.

\begin{table}[!t]
\centering
\caption{Results on CIFAR-100 dataset when applying LoCa on DKD.}
\begin{tabular}{lcc}
\toprule
\textbf{Teacher}   & \textbf{ResNet110} & \textbf{ResNet32x4} \\
\textbf{Student}  & \textbf{ResNet32} & \textbf{ShuffleNet-V2} \\
\midrule
DKD &   73.92       &   76.88      \\
\midrule
\textit{w/} LoCa ($\alpha$=0.95) &     73.96     &   \textbf{77.25}     \\
\textit{w/} LoCa ($\alpha$=0.98) &     73.95     &   77.09     \\
\textit{w/} LoCa ($\alpha$=0.99) &     \textbf{73.99}     &   77.18     \\
\bottomrule
\end{tabular}
\label{loca_dkd}
\end{table}

\subsection{Case Study}

We further showcase the text generation outputs of baselines and proposed LoCa.
The cases indicate that LoCa can effectively inherit the knowledge from the teacher model, such as the grammar information~(see Table \ref{case1}),
and avoiding the hallucinations~(see Table \ref{case2}).

Specifically, Table~\ref{case1} reports a case of concatenating the given sentences.
It shows that although the KD method connects the given sentences as instruction, it introduces extraneous punctuation and does not adhere fully to grammatical and syntactic norms like the extra \textbf{comma} and \textbf{colon} at the end. 
In contrast, LoCa perfectly replicates the teacher's output, effectively leveraging the distilled knowledge to seamlessly avoid errors. This superiority of LoCa is attributed to its ability to assimilate correct and comprehensive dark knowledge from the teacher output during training.

Table~\ref{case2} presents an example of hallucination, a common issue in large language model inference. 
Hallucination typically manifests as descriptions of related subjects rather than accurate responses to the standard prompts~(as Student answer in Table~\ref{case2}). 
In our analysis with LoCa, one potential reason is the student's unclear and imprecise understanding when transferring knowledge. 
LoCa addresses this by correcting the mis-instructions in the supervision information to be absolutely accurate, thereby mitigating the ambiguity and enhancing performance on the test set.

\begin{table}[!t]
\centering
\caption{One case from Dolly.
LoCa can avoid \textit{grammar errors} with the knowledge distilled from the teacher model, while KD fails.
}
\begin{tabular}{lp{0.35\textwidth}}
\toprule
\textbf{Instruction} & \textbf{You are given a list of sentences. Your task is to concatenate these sentences in the same order and print the result on a single line.} \\
\midrule
\textbf{Input} & Sentence1 = I'm happy today.,  \\&Sentence2 = This is a good day.,  \\&Sentence3 = I will succeed. \\
\midrule
\textbf{Grouth Truth} & I'm happy today. This is a good day. I will succeed. \\
\midrule
\textbf{Teacher} & I'm happy today. This is a good day. I will succeed. \\
\midrule
\textbf{Student} & I'm happiness today., This is a good day.", I will success \\
\midrule
\textbf{KD} & I'm happy today., This is a good day.", I will succeed. \\
\midrule
\textbf{LoCa(ours)} & I'm happy today. This is a good day. I will succeed. \\
\bottomrule
\end{tabular}

\label{case1}
\end{table}
\begin{table}[!t]
\centering
\caption{Another case from Dolly. 
The vanilla student model exists \textit{ hallucinations}.
LoCa can address this issue, but KD fails.
}
\begin{tabular}{lp{0.35\textwidth}}
\toprule
\textbf{Instruction} & \textbf{In this task, you are given a list of Amazon products. The task is to classify each product as "GOOD" if the majority of reviewers recommend the product or as "BAD" if the majority of reviewers do not recommend the product.} \\
\midrule
\textbf{Input} & Product1: 4 out of 5 stars, Product2: 3 out of 5 stars, Product3: 2 out of 5 stars. \\
\midrule
\textbf{Grouth Truth} & Product1: GOOD, Product2: BAD, Product3: BAD \\
\midrule
\textbf{Teacher} & Product1: GOOD, Product2: BAD, Product3: BAD \\
\midrule
\textbf{Student} & Product1: GOOD, Product3: BAD, Product2: GOOD, Product2: BAD Note: Three products (Product1, Product3, and Product2) have more than one review and are represented only once in the list. In this case, we classify them as the same label regardless of what majority rule (more positive reviews are SAFE) \\
\midrule
\textbf{KD} & Product1: Good. 20\% of the reviewers recommend this product. Product2: Bad. 80\% of the review do not recommend this product. Product3: Bad. 81\% of the review do not recommend this product. \\
\midrule
\textbf{LoCa (ours)} & Product1: GOOD, Product2: BAD, Product3: BAD \\
\bottomrule
\end{tabular}

\label{case2}
\end{table}

\section{Related Work}

\subsection{Logit-base Knowledge Distillation}

The logit-based knowledge distillation is originally proposed by Hinton~\citep{hinton2015distilling}, in which the student is trained to mimic the soft logit output of the teacher~\citep{wang2023riformer,wu2023edge}. 
Several methods have been developed to enhance its effectiveness.
SimKD~\citep{chen2022simKD} improves knowledge transfer by utilizing the teacher’s classifier within the student network.
DKD~\citep{zhao2022decoupled} proposes the decoupled knowledge distillation that divides logit knowledge into target knowledge and non-target knowledge.
NKD~\citep{yang2023knowledge} further proposes normalizing the non-target logits to equalize their sum. 
ATS~\citep{li2022asymmetric} decouples KD into three components, clarifying that knowledge transfer requires expanding the variance of incorrect category probabilities. It implements this improvement through dynamic temperature adjustments.
CTKD~\citep{li2023curriculum} improves knowledge distillation by assigning different temperatures to instances. 
Other works~\citep{zhu2022teach, qiu2022better, huang2022knowledge} refine the logit-based distillation paradigm to enhance the effectiveness when using stronger teacher models for distillation.
We find that logit-based distillation offers numerous advantages as a way to compress the model, making it a worthwhile focus for our further research.

\subsection{Logits Errorness}
SDD~\citep{luo2024scale} uncovers multi-label issues on ImageNet, using the logits map to partition sub-regions and enhance the information density of logits for more effective teaching to solve the mismatch between teacher's predictions to ground truth.
TIE-KD~\citep{Zhang2023Towards} focuses on discrepancies between teacher predictions and ground truth and emphasize the top-k predictions to enhance overall performance.
Our approach differs as it aims to ensure absolute consistency in the teacher's predictions while minimizing the gap.

\subsection{Probability Calibration}
Some methods require little extra or even no more time than training the model directly. For example, 
Label Smoothing~\citep{szegedy2016rethinking} sets the labels manually by distributing the same values to all non-target classes. 
Tf-KD~\citep{yuan2020revisiting} revisits KD via label smoothing, using a high temperature to generate the manual logit for distillation. 
LSKD~\citep{sun2024logit} explores the feasibility of varying temperature coefficients and achieves Logit standardization in knowledge distillation through Z transformations. 
Different from their works, we emphasize the use of simple linear transformations to preserve the relative proportions of non-target categories, thereby retaining the valuable dark knowledge within them.

\section{Conclusion}

This work revisits conventional logit-based distillation and reveals that the effectiveness of KD is limited by situations we term as \textbf{mis-instruction}, where the student model is misled when predictions based on teacher logits do not align with the ground truth labels.
To overcome this limitation, we propose a simple yet effective method called \textbf{Lo}git \textbf{Ca}libration (LoCa), which calibrate the supervision logits in cases of mis-instruction by decreasing non-target logits and enhancing target logits.
We establish knowledge-transferring pipelines for these mis-instruction outputs to transfer accurate target information and other useful relation information.
This approach ensures that the adjusted target logits are absolutely correct, while preserving the relative proportions among non-target logits to maintain dark knowledge. Our proposed LoCa method does not require any additional parameters. 
Extensive experiments on several benchmark datasets, including image classification and text generation tasks, demonstrate the effectiveness of LoCa across a range of teacher-student pairs.

\section{Ethics Statement}

This research utilized publicly available datasets in the fields of Computer Vision (CV) and Natural Language Processing (NLP), which do not involve sensitive personal information or human subjects, thus not requiring specific ethical approvals. We adhere to ethical guidelines respecting privacy and intellectual property rights of the data sources. Our project develops tools intended to support, not substitute, the anticipatory processes in AI technologies. We emphasize the non-commercial use of the datasets and fine-tuned models, and the adherence to fair use in data sourcing. Conflicts of interest do not influence this research, which ensures its integrity and the reliability of its results.

\begin{ack}
This work was partially supported by the National Natural Science Foundation of China (Grant No. 61991451) and the Shenzhen Science and Technology Program (JCYJ20220818101001004).
\end{ack}

\clearpage


\begin{thebibliography}{44}
\providecommand{\natexlab}[1]{#1}
\providecommand{\url}[1]{\texttt{#1}}
\expandafter\ifx\csname urlstyle\endcsname\relax
  \providecommand{\doi}[1]{doi: #1}\else
  \providecommand{\doi}{doi: \begingroup \urlstyle{rm}\Url}\fi

\bibitem[Chandrasegaran et~al.(2022)Chandrasegaran, Tran, Zhao, and Cheung]{chandrasegaran2022revisiting}
K.~Chandrasegaran, N.-T. Tran, Y.~Zhao, and N.-M. Cheung.
\newblock Revisiting label smoothing and knowledge distillation compatibility: What was missing?
\newblock In \emph{International Conference on Machine Learning}, pages 2890--2916, 2022.

\bibitem[Chen et~al.(2022)Chen, Mei, Zhang, Wang, Feng, and Chen]{chen2022simKD}
D.~Chen, J.-P. Mei, H.~Zhang, C.~Wang, Y.~Feng, and C.~Chen.
\newblock Knowledge distillation with the reused teacher classifier.
\newblock In \emph{Proceedings of the IEEE/CVF Conference on Computer Vision and Pattern Recognition}, pages 11933--11942, 2022.

\bibitem[Chen et~al.(2021)Chen, Liu, Zhao, and Jia]{chen2021distilling}
P.~Chen, S.~Liu, H.~Zhao, and J.~Jia.
\newblock Distilling knowledge via knowledge review.
\newblock In \emph{Proceedings of the IEEE/CVF Conference on Computer Vision and Pattern Recognition}, pages 5008--5017, 2021.

\bibitem[Devlin et~al.(2019)Devlin, Chang, Lee, and Toutanova]{DBLP:conf/naacl/DevlinCLT19}
J.~Devlin, M.~Chang, K.~Lee, and K.~Toutanova.
\newblock {BERT:} pre-training of deep bidirectional transformers for language understanding.
\newblock In \emph{Proceedings of the 2019 Conference of the North American Chapter of the Association for Computational Linguistics: Human Language Technologies}, volume 1 (Long and Short Papers), pages 4171--4186, 2019.

\bibitem[Gu et~al.(2023)Gu, Dong, Wei, and Huang]{DBLP:journals/corr/abs-2306-08543}
Y.~Gu, L.~Dong, F.~Wei, and M.~Huang.
\newblock Knowledge distillation of large language models.
\newblock \emph{arXiv preprint arXiv:2306.08543}, 2023.

\bibitem[He et~al.(2016)He, Zhang, Ren, and Sun]{he2016deep}
K.~He, X.~Zhang, S.~Ren, and J.~Sun.
\newblock Deep residual learning for image recognition.
\newblock In \emph{Proceedings of the IEEE Conference on Computer Vision and Pattern Recognition}, pages 770--778, 2016.

\bibitem[He et~al.(2017)He, Gkioxari, Doll{\'a}r, and Girshick]{mask}
K.~He, G.~Gkioxari, P.~Doll{\'a}r, and R.~Girshick.
\newblock Mask {R}-{C}{N}{N}.
\newblock In \emph{Proceedings of the IEEE/CVF International Conference on Computer Vision}, pages 2961--2969, 2017.

\bibitem[Hinton et~al.(2015)Hinton, Vinyals, and Dean]{hinton2015distilling}
G.~Hinton, O.~Vinyals, and J.~Dean.
\newblock Distilling the knowledge in a neural network.
\newblock \emph{arXiv preprint arXiv:1503.02531}, 2015.

\bibitem[Honovich et~al.(2023)Honovich, Scialom, Levy, and Schick]{honovich-etal-2023-unnatural}
O.~Honovich, T.~Scialom, O.~Levy, and T.~Schick.
\newblock Unnatural instructions: Tuning language models with (almost) no human labor.
\newblock In \emph{Proceedings of the 61st Annual Meeting of the Association for Computational Linguistics (Volume 1: Long Papers)}, pages 14409--14428, 2023.

\bibitem[Howard(2017)]{howard2017mobilenets}
A.~G. Howard.
\newblock Mobilenets: Efficient convolutional neural networks for mobile vision applications.
\newblock \emph{arXiv preprint arXiv:1704.04861}, 2017.

\bibitem[Hu et~al.(2018)Hu, Shen, and Sun]{senet}
J.~Hu, L.~Shen, and G.~Sun.
\newblock Squeeze-and-excitation networks.
\newblock In \emph{Proceedings of the IEEE Conference on Computer Vision and Pattern Recognition}, pages 7132--7141, 2018.

\bibitem[Huang et~al.(2022)Huang, You, Wang, Qian, and Xu]{huang2022knowledge}
T.~Huang, S.~You, F.~Wang, C.~Qian, and C.~Xu.
\newblock Knowledge distillation from a stronger teacher.
\newblock \emph{Advances in Neural Information Processing Systems}, 35:\penalty0 33716--33727, 2022.

\bibitem[Kim and Rush(2016)]{kim2016sequence}
Y.~Kim and A.~M. Rush.
\newblock Sequence-level knowledge distillation.
\newblock \emph{arXiv preprint arXiv:1606.07947}, 2016.

\bibitem[Krizhevsky et~al.(2009)Krizhevsky, Hinton, et~al.]{cifar}
A.~Krizhevsky, G.~Hinton, et~al.
\newblock Learning multiple layers of features from tiny images.
\newblock 2009.

\bibitem[Lao et~al.(2023{\natexlab{a}})Lao, Song, Liu, Liu, and Yang]{lao2023masked}
S.~Lao, G.~Song, B.~Liu, Y.~Liu, and Y.~Yang.
\newblock Masked autoencoders are stronger knowledge distillers.
\newblock In \emph{Proceedings of the IEEE/CVF International Conference on Computer Vision}, pages 6384--6393, 2023{\natexlab{a}}.

\bibitem[Lao et~al.(2023{\natexlab{b}})Lao, Song, Liu, Liu, and Yang]{lao2023unikd}
S.~Lao, G.~Song, B.~Liu, Y.~Liu, and Y.~Yang.
\newblock Unikd: Universal knowledge distillation for mimicking homogeneous or heterogeneous object detectors.
\newblock In \emph{Proceedings of the IEEE/CVF International Conference on Computer Vision}, pages 6362--6372, 2023{\natexlab{b}}.

\bibitem[Li et~al.(2022)Li, Fan, Song, Li, Yunfeng, Zhan, et~al.]{li2022asymmetric}
X.-C. Li, W.-S. Fan, S.~Song, Y.~Li, S.~Yunfeng, D.-C. Zhan, et~al.
\newblock Asymmetric temperature scaling makes larger networks teach well again.
\newblock \emph{Advances in Neural Information Processing Systems}, 35:\penalty0 3830--3842, 2022.

\bibitem[Li et~al.(2023)Li, Li, Yang, Zhao, Song, Luo, Li, and Yang]{li2023curriculum}
Z.~Li, X.~Li, L.~Yang, B.~Zhao, R.~Song, L.~Luo, J.~Li, and J.~Yang.
\newblock Curriculum temperature for knowledge distillation.
\newblock In \emph{Proceedings of the AAAI Conference on Artificial Intelligence}, volume~37, pages 1504--1512, 2023.

\bibitem[Liu et~al.(2019)Liu, Ott, Goyal, et~al.]{DBLP:journals/corr/abs-1907-11692}
Y.~Liu, M.~Ott, N.~Goyal, et~al.
\newblock Roberta: A robustly optimized bert pretraining approach.
\newblock \emph{arXiv preprint arXiv:1907.11692}, 2019.

\bibitem[Long et~al.(2015)Long, Shelhamer, and Darrell]{shelhamer2016fully}
J.~Long, E.~Shelhamer, and T.~Darrell.
\newblock Fully convolutional networks for semantic segmentation.
\newblock In \emph{Proceedings of the IEEE Conference on Computer Vision and Pattern Recognition}, pages 3431--3440, 2015.

\bibitem[Luo(2024)]{luo2024scale}
S.~W. C. L.~Y. Luo.
\newblock Scale decoupled distillation.
\newblock \emph{arXiv preprint arXiv:2403.13512}, 2024.

\bibitem[Ma et~al.(2018)Ma, Zhang, Zheng, and Sun]{shufflenetv2}
N.~Ma, X.~Zhang, H.-T. Zheng, and J.~Sun.
\newblock Shufflenet v2: Practical guidelines for efficient cnn architecture design.
\newblock In \emph{Proceedings of the European Conference on Computer Vision}, pages 116--131, 2018.

\bibitem[Qiu et~al.(2022)Qiu, Ma, Yang, Liu, Hou, Yi, and Ouyang]{qiu2022better}
Z.~Qiu, X.~Ma, K.~Yang, C.~Liu, J.~Hou, S.~Yi, and W.~Ouyang.
\newblock Better teacher better student: Dynamic prior knowledge for knowledge distillation.
\newblock \emph{arXiv preprint arXiv:2206.06067}, 2022.

\bibitem[Ren et~al.(2016)Ren, He, Girshick, and Sun]{faster_rcnn}
S.~Ren, K.~He, R.~Girshick, and J.~Sun.
\newblock Faster r-cnn: Towards real-time object detection with region proposal networks.
\newblock \emph{IEEE Transactions on Pattern Analysis and Machine Intelligence}, 39\penalty0 (6):\penalty0 1137--1149, 2016.

\bibitem[Russakovsky et~al.(2015)Russakovsky, Deng, Su, Krause, Satheesh, Ma, Huang, Karpathy, Khosla, Bernstein, et~al.]{imagenet}
O.~Russakovsky, J.~Deng, H.~Su, J.~Krause, S.~Satheesh, S.~Ma, Z.~Huang, A.~Karpathy, A.~Khosla, M.~Bernstein, et~al.
\newblock Imagenet large scale visual recognition challenge.
\newblock \emph{International Journal of Computer Vision}, 115:\penalty0 211--252, 2015.

\bibitem[Sandler et~al.(2018)Sandler, Howard, Zhu, Zhmoginov, and Chen]{sandler2018mobilenetv2}
M.~Sandler, A.~Howard, M.~Zhu, A.~Zhmoginov, and L.-C. Chen.
\newblock Mobilenetv2: Inverted residuals and linear bottlenecks.
\newblock In \emph{Proceedings of the IEEE Conference on Computer Vision and Pattern Recognition}, pages 4510--4520, 2018.

\bibitem[Sun et~al.(2024)Sun, Ren, Li, Wang, and Cao]{sun2024logit}
S.~Sun, W.~Ren, J.~Li, R.~Wang, and X.~Cao.
\newblock Logit standardization in knowledge distillation.
\newblock In \emph{Proceedings of the IEEE/CVF Conference on Computer Vision and Pattern Recognition}, pages 15731--15740, 2024.

\bibitem[Sutskever et~al.(2013)Sutskever, Martens, Dahl, and Hinton]{sutskever2013importance}
I.~Sutskever, J.~Martens, G.~Dahl, and G.~Hinton.
\newblock On the importance of initialization and momentum in deep learning.
\newblock In \emph{International Conference on Machine Learning}, pages 1139--1147, 2013.

\bibitem[Szegedy et~al.(2016)Szegedy, Vanhoucke, Ioffe, Shlens, and Wojna]{szegedy2016rethinking}
C.~Szegedy, V.~Vanhoucke, S.~Ioffe, J.~Shlens, and Z.~Wojna.
\newblock Rethinking the inception architecture for computer vision.
\newblock In \emph{Proceedings of the IEEE Conference on Computer Vision and Pattern Recognition}, pages 2818--2826, 2016.

\bibitem[Touvron et~al.(2023)Touvron, Lavril, Izacard, Martinet, Lachaux, Lacroix, Rozi{\`e}re, Goyal, Hambro, Azhar, et~al.]{DBLP:journals/corr/abs-2302-13971}
H.~Touvron, T.~Lavril, G.~Izacard, X.~Martinet, M.-A. Lachaux, T.~Lacroix, B.~Rozi{\`e}re, N.~Goyal, E.~Hambro, F.~Azhar, et~al.
\newblock Llama: Open and efficient foundation language models.
\newblock \emph{arXiv preprint arXiv:2302.13971}, 2023.

\bibitem[Wang et~al.(2023)Wang, Zhang, Liu, Wu, Yang, Liu, Chen, Luo, and Lin]{wang2023riformer}
J.~Wang, S.~Zhang, Y.~Liu, T.~Wu, Y.~Yang, X.~Liu, K.~Chen, P.~Luo, and D.~Lin.
\newblock Riformer: Keep your vision backbone effective but removing token mixer.
\newblock In \emph{Proceedings of the IEEE/CVF Conference on Computer Vision and Pattern Recognition}, pages 14443--14452, 2023.

\bibitem[Wang et~al.(2022)Wang, Mishra, Alipoormolabashi, et~al.]{wang-etal-2022-super}
Y.~Wang, S.~Mishra, P.~Alipoormolabashi, et~al.
\newblock Super-{N}atural{I}nstructions: Generalization via declarative instructions on 1600+ {NLP} tasks.
\newblock In \emph{Proceedings of the 2022 Conference on Empirical Methods in Natural Language Processing}, pages 5085--5109, 2022.

\bibitem[Wu et~al.(2023)Wu, Zhao, Wang, Bai, Wang, Wong, and Yang]{wu2023edge}
T.~Wu, Z.~Zhao, J.~Wang, X.~Bai, L.~Wang, N.~Wong, and Y.~Yang.
\newblock Edge-free but structure-aware: Prototype-guided knowledge distillation from gnns to mlps.
\newblock \emph{arXiv preprint arXiv:2303.13763}, 2023.

\bibitem[Wu et~al.(2024{\natexlab{a}})Wu, Hou, Lao, Li, Wong, Zhao, and Yang]{wu2024weight}
T.~Wu, C.~Hou, S.~Lao, J.~Li, N.~Wong, Z.~Zhao, and Y.~Yang.
\newblock Weight-inherited distillation for task-agnostic bert compression.
\newblock In \emph{Findings of the Association for Computational Linguistics: NAACL 2024}, pages 13--28, 2024{\natexlab{a}}.

\bibitem[Wu et~al.(2024{\natexlab{b}})Wu, Tao, Wang, Zhao, and Wong]{wu2024rethinking}
T.~Wu, C.~Tao, J.~Wang, Z.~Zhao, and N.~Wong.
\newblock Rethinking kullback-leibler divergence in knowledge distillation for large language models.
\newblock \emph{arXiv preprint arXiv:2404.02657}, 2024{\natexlab{b}}.

\bibitem[Yang et~al.(2023)Yang, Zeng, Li, Zhang, Yuan, and Li]{yang2023knowledge}
Z.~Yang, A.~Zeng, Z.~Li, T.~Zhang, C.~Yuan, and Y.~Li.
\newblock From knowledge distillation to self-knowledge distillation: A unified approach with normalized loss and customized soft labels.
\newblock In \emph{Proceedings of the IEEE/CVF International Conference on Computer Vision}, pages 17185--17194, 2023.

\bibitem[Yuan et~al.(2020)Yuan, Tay, Li, Wang, and Feng]{yuan2020revisiting}
L.~Yuan, F.~E. Tay, G.~Li, T.~Wang, and J.~Feng.
\newblock Revisiting knowledge distillation via label smoothing regularization.
\newblock In \emph{Proceedings of the IEEE/CVF Conference on Computer Vision and Pattern Recognition}, pages 3903--3911, 2020.

\bibitem[Zagoruyko and Komodakis(2016)]{zagoruyko2016wide}
S.~Zagoruyko and N.~Komodakis.
\newblock Wide residual networks.
\newblock \emph{arXiv preprint arXiv:1605.07146}, 2016.

\bibitem[Zhang et~al.(2024)Zhang, Zeng, Wang, and Lu]{zhang2024tinyllama}
P.~Zhang, G.~Zeng, T.~Wang, and W.~Lu.
\newblock Tinyllama: An open-source small language model.
\newblock \emph{arXiv preprint arXiv:2401.02385}, 2024.

\bibitem[Zhang et~al.(2023)Zhang, Liang, Wang, Han, Liu, Xu, and Chen]{Zhang2023Towards}
S.~Zhang, Y.~Liang, S.~Wang, W.~Han, J.~Liu, J.~Xu, and Y.~Chen.
\newblock Towards understanding and improving knowledge distillation for neural machine translation.
\newblock In \emph{Annual Meeting of the Association for Computational Linguistics}, pages 8062--8079, 2023.

\bibitem[Zhang et~al.(2018)Zhang, Zhou, Lin, and Sun]{zhang2018shufflenet}
X.~Zhang, X.~Zhou, M.~Lin, and J.~Sun.
\newblock Shufflenet: An extremely efficient convolutional neural network for mobile devices.
\newblock In \emph{Proceedings of the IEEE Conference on Computer Vision and Pattern Recognition}, pages 6848--6856, 2018.

\bibitem[Zhao et~al.(2022)Zhao, Cui, Song, Qiu, and Liang]{zhao2022decoupled}
B.~Zhao, Q.~Cui, R.~Song, Y.~Qiu, and J.~Liang.
\newblock Decoupled knowledge distillation.
\newblock In \emph{Proceedings of the IEEE/CVF Conference on Computer Vision and Pattern Recognition}, pages 11953--11962, 2022.

\bibitem[Zhao et~al.(2017)Zhao, Shi, Qi, Wang, and Jia]{pspnet}
H.~Zhao, J.~Shi, X.~Qi, X.~Wang, and J.~Jia.
\newblock Pyramid scene parsing network.
\newblock In \emph{Proceedings of the IEEE Conference on Computer Vision and Pattern Recognition}, pages 2881--2890, 2017.

\bibitem[Zhu et~al.(2022)Zhu, Liu, Xu, Liu, Meng, Wang, Ou, and Tang]{zhu2022teach}
Y.~Zhu, N.~Liu, Z.~Xu, X.~Liu, W.~Meng, L.~Wang, Z.~Ou, and J.~Tang.
\newblock Teach less, learn more: On the undistillable classes in knowledge distillation.
\newblock \emph{Advances in Neural Information Processing Systems}, 35:\penalty0 32011--32024, 2022.

\end{thebibliography}


\end{document}